\documentclass[letterpaper, 10 pt, conference]{ieeeconf}  

\IEEEoverridecommandlockouts                              
\usepackage{graphicx}   
\usepackage{amsmath}
\usepackage{amssymb}
\usepackage{mathtools}
\usepackage{booktabs} 
\usepackage{hyperref}
\usepackage{caption}
\usepackage{subcaption}
\usepackage[export]{adjustbox}
\usepackage{soul}
\usepackage{makecell}
\usepackage{xcolor}
\usepackage{balance}
\title{\LARGE \bf
Biomechanically consistent real-time action recognition\\for human-robot interaction
}

\author{Wanchen Li$^{1}$, Kahina Chalabi$^{2}$, Maxime Sabbah$^{2}$, Thomas Bousquet$^{1,2}$,  \\ Robin Passama$^{1}$, Sofiane Ramdani$^{1}$, Andrea Cherubini$^{3}$, Vincent Bonnet$^{2,4}$
\thanks{${^1}$ LIRMM, University of Montpellier,
Montpellier, France.}
\thanks{${^2}$ LAAS-CNRS, Université Paul Sabatier, CNRS, Toulouse, France.}
\thanks{${^3}$ Nantes Université, École Centrale Nantes, CNRS, LS2N, UMR 6004, 1, rue de la Noe, 44321 Nantes, France.}
\thanks{${^4}$ IPAL, CNRS, Singapore.}}

\begin{document}

\maketitle
\thispagestyle{empty}
\pagestyle{empty}
\begin{abstract} 
This paper presents a novel framework for real-time human action recognition in industrial contexts, using standard 2D cameras. We introduce a complete pipeline for robust and real-time estimation of human joint kinematics, input to a temporally smoothed Transformer-based network, for action recognition. We rely on a new dataset including 11 subjects performing various actions, to evaluate our approach. Unlike most of the literature that relies on joint center positions (JCP) and is offline, ours uses  biomechanical prior, eg. joint angles, for fast and robust real-time recognition. Besides, joint angles make the proposed method agnostic to sensor and subject poses as well as to anthropometric differences, and ensure robustness across environments and subjects. Our proposed learning model outperforms the best baseline model, running also in real-time, along various metrics. It achieves 88\% accuracy and shows great generalization ability, for subjects not facing the cameras. Finally, we demonstrate the robustness and usefulness of our technique, through an online interaction experiment, with a simulated robot controlled in real-time via the recognized actions.
\end{abstract}

\section{Introduction}

Online human action recognition (OHAR) is a critical challenge in computer vision, collaborative robotics and human-machine interaction. It is much more difficult than offline human action recognition and it is particularly important in industrial robotics, and factory automation. Traditionally, human action recognition algorithms are developed upon trimmed datasets, in which different actions are already segmented into different sequences, the post-processing step is only to recognize the action inside specific sequence. Obviously, this approach is unsuitable for real-time robot control, where actions happen continuously without a manual segmentation step. Surprisingly, few studies are shifting to online action recognition \cite{de2016online, NEURIPS2021_08b255a5, 11091374}, where actions are classified within a sliding window, relying only on recent past information. Such frameworks are more suited for robotics, to promptly respond to human actions. 

Robotic systems can perceive human actions thanks to multiple modalities. The survey \cite{ajoudani2018progress} outlined key perception methods, including vision, audio-based interfaces, force sensors, and bio-signals, e.g., electromyography or electroencephalography. While many systems integrate these into hybrid setups, cumbersome devices are impractical on factory floors. Vision-based perception interfaces offer the most feasible solution in the factory settings. Human body gestures naturally communicate abundant information in industrial contexts. Several approaches \cite{barattini2012gesture, mazhar2019real} demonstrate the effectiveness of hand gestures, head orientations, and task-based gestures in conveying orders and needs from human workers.

Our work on human action recognition is pose-based. For pose estimation, numerous learning-based algorithms leveraging convolutional neural networks or Transformers — such as OpenPose \cite{cao2019openpose}, MMPose \cite{mmpose2020}, YOLOv7 \cite{yolov7}, and ViTPose \cite{vitpose}—have reached a high level of maturity. These markerless pose estimation algorithms extract skeletons from images or video frames and identify 2D joint centers, called \textit{keypoints}, without camera calibration. While being real-time, these methods do not guarantee sufficient accuracy for realistic body structures. Indeed, the keypoint outputs are often noisy, limiting their use in robotics. The low number of keypoints also restricts the estimation to simple, coarse-grained movements. The work \cite{li2021hybrik} attempted to extend keypoint estimation to joint angle estimation for improving pose estimation, using a purely data-based recursive approach. Researchers from Stanford \cite{opencap} also proposed a dual camera setup and trained a Long Short-Term Memory (LSTM) network for keypoint data augmentation and filtering, leading to precise joint angle estimation with an accuracy below 5 degrees for rehabilitation tasks, but not in real-time. In our work, we integrate a biomechanically consistent model prior allowing to regulate the joint angles based on noisy real-time keypoints. Joint angle estimation is performed in a one-shot manner, independent of the perception viewpoints, and robust to individual anthropomorphic differences, making it suitable for online action recognition. 

We hereby present a real-time human action recognition pipeline combining biomechanical modeling and machine learning. It is based on:

\begin{itemize} 
\item a robust real-time method for joint angle estimation based on a biomechanical model computing from the estimated skeleton keypoints (Section \ref{pipeIK}), 
\item a Transformer model with a temporal causal mask trained by a combined loss function, (Section \ref{transformer}).
\end{itemize}

The main contributions of our work with respect to the state-of-the-art are the following:
\begin{itemize}
    \item We use a biomechanical model prior to regulate skeleton keypoint estimation, obtaining frame-wise joint angles for human action recognition. 

    \item We employ joint angles as a compact and expressive representation of human motion. Leveraging a new dataset, HUMAR-24, introduced in Section~\ref{dataset}, we show that joint angle kinematics are effective for fine-grained action recognition and offer improved generalization across different viewpoints (Section~\ref{Deep_exp}). The use of joint angles also allows distinguishing upper-body action and lower-body action.

    \item
    The proposed model adheres to the International Society of Biomechanics conventions, ensuring biomechanical consistency and facilitating its use in ergonomics studies.

    \item 
    A simulation experiment demonstrates the applicability of this pipeline for real-time robot control (Section~\ref{robotexp}).
\end{itemize}

\section{Related works}

Numerous existing action recognition methods are designed for short, segmented videos, where each clip contains a single, well-defined action. Deep learning approaches such as 3D convolutional neural networks (CNNs) (e.g., 3D CNN \cite{ji20123d}, C3D \cite{hara2017learning}, LTC \cite{varol2017long}), recurrent networks like LSTMs \cite{liu2017global, zhang2019view, du2015hierarchical}, and more recently, graph convolutional networks (GCNs) \cite{chen2021channel, song2022constructing}, have been developed under these assumptions. These models are typically trained and evaluated on trimmed action recognition benchmarks such as MSR Action 3D, NTU RGB+D, KTH, and Human3.6M \cite{song2021human}, where action boundaries are clearly labeled.
While effective in controlled settings, these methods are ill-suited for real-world robotic applications, where actions occur continuously over time and must be recognized online from unsegmented input streams. Some recent models claim real-time inference capabilities \cite{kopuklu2019you, sanchez20223dfcnn}, but they are still trained on trimmed datasets; this limits their adaptability to dynamic, real-world scenarios. 


 Recent works on OHAR trained on untrimmed datasets have primarily relied on image-based features, resulting in computationally heavy models with large parameter counts \cite{NEURIPS2021_08b255a5, KIM2021107954}. However, methods that operate directly on RGB images are inherently sensitive to visual variations such as background clutter and viewpoint changes. In contrast, pose-based approaches leverage compact and structured skeleton representations, offering greater efficiency and improved generalization across different scenes and viewpoints. Given our focus on efficient and viewpoint-agnostic action recognition, the comparison in this study is restricted to pose-based methods. Within this context, to the best of our knowledge, the work by Mazzia et al. \cite{mazzia2022action} represents the only directly comparable baseline. This state-of-the-art study \cite{mazzia2022action} uses a Transformer encoder model to process skeleton-based features in Cartesian space, avoiding image-based embeddings. However, a Transformer encoder that relies solely on 2D keypoints (more formally referred to as joint center positions, JCPs) as input may still produce unstable recognition results, particularly in dynamic robotic contexts. In fact, joint positions are defined in a global reference frame, and are therefore sensitive to changes of the camera viewpoint. In many robotic applications, the perception device (e.g., a camera mounted on a mobile robot or manipulator) is not stationary: its orientations and/or positions vary over time. Consequently, a model trained on data from a fixed viewpoint may struggle to generalize under such conditions.

To mitigate this issue, we use joint angle kinematics directly as model input. This representation is inherently invariant to the camera viewpoints, since it encodes a pose in terms of relative joint rotations rather than absolute positions. We demonstrate that this input modality significantly outperforms JCP-based representations, when evaluated on the Transformer architecture and on a test set that includes viewpoint variations.

\section{Methods}
In this section, we present our pipeline \ref{fig:pipeIK} for real-time action recognition using 1) a \textit{real-time human joint kinematics estimation} over 2) our \textit{HUMAR-2024 continuous action dataset} that serves as input to 3) a \textit{temporal-attentive Transformer-based network}. 

\subsection{Real-time estimation of human joint kinematics}
\label{pipeIK}
\begin{figure}[t]
  \centering
  \includegraphics[width=1.05\columnwidth,clip,trim=0 0 0 0]{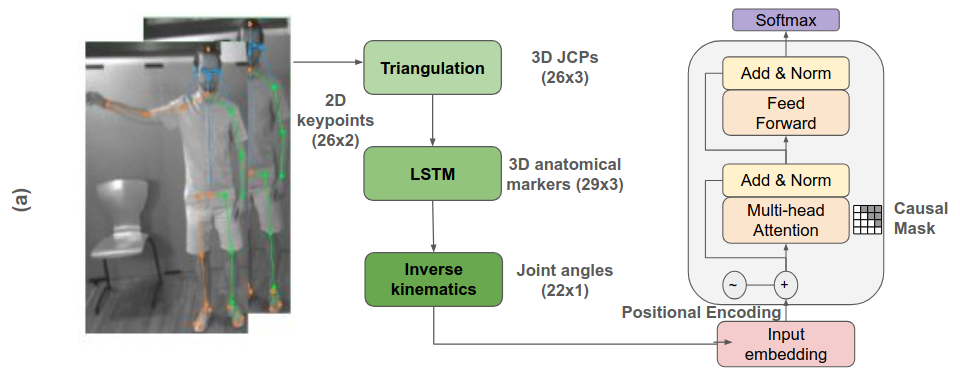}
  \includegraphics[width=1\columnwidth,clip,trim=0 0 0 0]{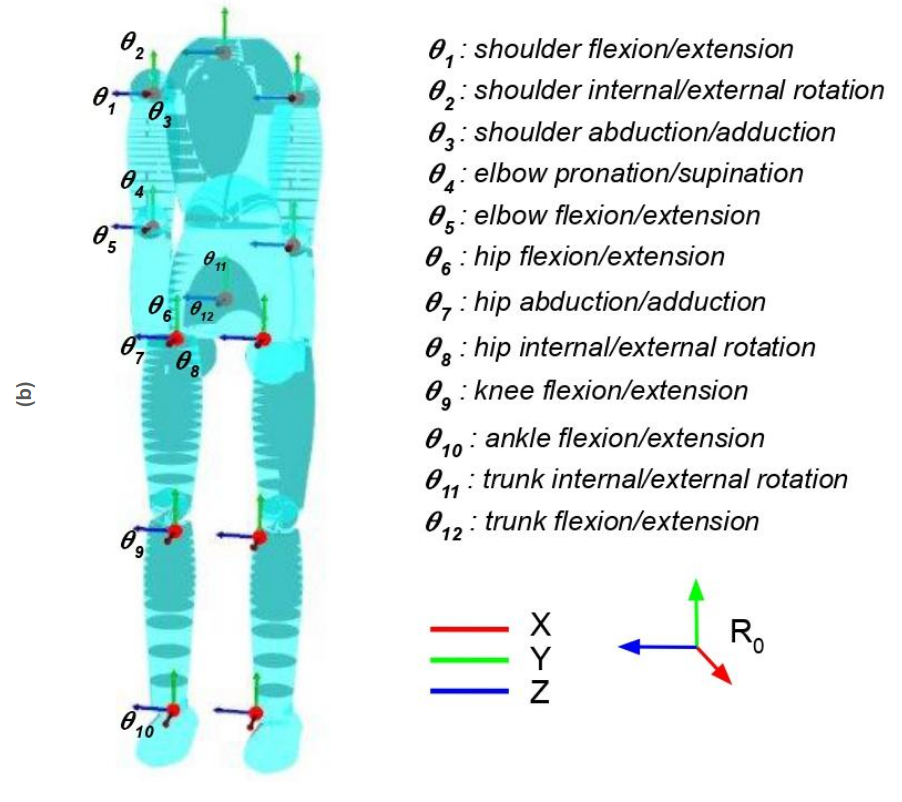}
  \setlength{\belowcaptionskip}{-5pt}
  \caption{Representation of the real-time inverse kinematics pipeline 
   \newline (a). 22 degrees-of-freedom biomechanical model of the human body used in this study \newline (b). 
   For readability, only the joint angles of the right side of the body are explicated in the figure.}
  \label{fig:pipeIK}
\end{figure}

Accurately estimating human joint kinematics in real-time using affordable cameras remains a challenging task \cite{adjel_biorob_2023}. As shown in Fig. \ref{fig:pipeIK}.a, our pipeline uses RTMPose, a state-of-the-art human pose estimator, to extract 2D keypoint coordinates \cite{Jiang2023RTMPoseRM} based on data captured from two global shutter cameras, followed by triangulation, combined with an LSTM network for marker augmetation, to estimate 3D joint centers and virtual markers \cite{opencap_bench}. Then, a biomechanical model coupled with fast Quadratic Programming optimization is used to perform inverse kinematics from the augmented and filtered 3D data.

Recently, the RTMPose-t (tiny) model \cite{Jiang2023RTMPoseRM} was proposed for its fast execution time. That model relies on a lightweight convolutional network optimized for efficiency. Instead of traditional heatmaps, it adopts a classification-based approach, treating the 26 joint centers x and y coordinates as separate tasks, hence reducing computation while maintaining accuracy.

The predicted 2D keypoint positions from RTMPose are firstly fused by triangulation \cite{abdel2015direct} on two calibrated\cite{zhang2002flexible} cameras, to obtain 3D joint center positions. Nevertheless, the 26 3D joint center positions were not sufficient to fully capture the complexity of human joint kinematics. For example, if only the wrist and elbow position are known, it is not possible to determine the shoulder internal-external rotation. To address this limitation, we employed a LSTM-based data augmenter \cite{opencap} to predict the 3D positions of a more detailed set of $N_m=29$ anatomical markers. This marker set allows a much better representation of complex movements than the original set of keypoints. 

The biomechanical model was calibrated using the anatomical markers' positions obtained during the first pose when the subject was standing and facing the camera. Then, a classic real-time Quadratic Programming inverse kinematics calculation \cite{escande2014hierarchical} was performed to retrieve the vector $\mathbf{q}$ of the 22 joint angles that best fit given 3D markers position measurements. 


As illustrated in Fig. \ref{fig:pipeIK}.b, the retained biomechanical model followed the International Society of Biomechanics guidelines to define segment frame orientations and joint rotation sequences \cite{wu_upper, wu_lower}, ensuring future applicability for ergonomics studies. 

The total execution time of the proposed pipeline is approximately 47.5 ms per frame on a  computer equipped with an NVIDIA RTX 2070 GPU (8 GB memory) and a 12-core Intel(R) Core(TM) i7-9750H CPU @ 2.60GHz. This time includes approximately 27ms for RTMPose processing of two simultaneous images, 5.5ms for triangulation and LSTM-based temporal filtering, and 15ms for solving the inverse kinematics problem. 
Note that this real-time markerless inverse kinematics pipeline is available as an open-source toolbox at \url{https://gitlab.laas.fr/msabbah/rt-cosmik}
. We are confident that with a more recent hardware setup, including a larger GPU, this real-time markerless inverse kinematics pipeline could run beyond 50Hz.

\subsection{Dataset}
\label{dataset}
Although there are many benchmark datasets for the action recognition task, most of them provide only image input acquired from a single camera. As discussed previously, today, monocular pose estimation methods cannot provide accurate and reliable 3D joint angles estimation. Thus, using two global shutter HD grayscale cameras, we created a custom dataset, namely HUMAR-2024. The camera frame rate was set to 10 Hz to allow sufficient time for processing, given the current hardware limitations.

Three female and eight male participants ($26\pm2$years, $1.72\pm0.09$m, $70.21\pm15.88$kg) took part in the experiment after providing written informed consent. Experiments were conducted with the national ethics review board of the Université Fédérale de Toulouse (IRB 00011835-2024-0910-855).
Each participant performed 20 actions, during a 3-minute continuous trial. These actions were selected and combined from four categories — \textit{motions}, \textit{transitory motions}, \textit{ordering actions}, and \textit{background actions} — resulting in a total of 17 distinct labels, detailed in Table \ref{tab:action_table}. Each participant completed 10 trials, which led to a dataset of 2,200 action instances on the 11 subjects. Given the limited data available per subject, a strict subject-exclusive split would markedly reduce training diversity and destabilize optimization. Therefore, a sequence-wise partitioning scheme is adopted, using 9 sequences from all 11 subjects for training and validation, and the remaining sequence for testing. As each subject’s sequences consist of randomly composed and heterogeneous actions, this strategy preserves inter-subject variability while evaluating on unseen temporal compositions, yielding a reliable assessment of generalization under data-scarce conditions.
The randomization of actions are done both in type and temporal order, to ensure balanced representation; each action label was encountered the same number of times across participants.
Motions refer to steady-state behaviors such as walking or standing.
During these motions, participants could simultaneously perform ordering actions (e.g., signaling a robot with the left or right arm) or background actions (e.g., crossing arms, doing nothing).
Transitory motions occurred between two steady states and were individually labeled — for instance, transitioning from a sitting to a standing position required a "standing up" label.
This dataset, albeit simplified, was designed to mirror real-world human-robot interaction scenarios, where human operators naturally transition between postures while executing tasks, offering a challenging and realistic benchmark for continuous action recognition.

\begin{table}[ht]
\centering
\resizebox{0.485\textwidth}{!}{%
\begin{tabular}{|l|l|l|l|}
\hline
\textbf{Motions} & \textbf{Transitory motions} & \textbf{Ordering actions} & \textbf{Background actions} \\ \hline
Standing \hfill 1& Standing up \hfill 5& Two arms picking \hfill 8& Examples: \hfill \\ 
Walking \hfill 2& Sitting down \hfill 6& Right arm picking \hfill 9& left hand on hip \hfill 
\\ 
Sitting \hfill 3& Squatting down \hfill 7& Left arm picking \hfill 10& right hand on hip, \hfill \\ 
Squatting \hfill 4& & Idle \hfill 11& cross arms, \hfill \\ 
& & Two thumbs up \hfill 12& left hand under chin, \hfill \\ 
& & Come sign by right arm \hfill 13& right hand under chin \hfill \\ 
& & Come sign by left arm \hfill 14& hands on hips \\ 
& & Come sign by two arms \hfill 15& etc \hfill 17\\ 
& & Stop sign \hfill 16& \\ 
\hline
\end{tabular}%
}
\caption{Motions, Transitory motions, Ordering actions, and Background actions. The numbers indicate the label count.}
\label{tab:action_table}
\end{table}

Participants followed prompts displayed on a screen and responded to action cues appearing at randomized intervals ranging from 5 to 15 seconds, each preceded by a countdown timer. Observations indicated that participants typically adjusted their posture within one second of the cue’s appearance. Based on this behavior, we determined that a 2-second buffer window was sufficient to capture the full duration of all dynamic actions.
To enable 3D pose estimation, we recorded videos using two cameras placed 30 cm apart, with parallel optical axes. This configuration ensured that both views captured the scene simultaneously, allowing for accurate triangulation of joint positions.
We split the dataset into training (70\%), validation (20\%), and test (10\%) sets. For each time step, we applied the markerless inverse kinematics pipeline described in Section \ref{pipeIK} to extract the joint angle vector $\mathbf{q}$, comprising 22 joint angles. This resulted in a dataset with 22 input features and 17 action labels per sample.

\subsection{Transformer network}
\label{transformer}
\begin{figure}[t]
  \centering
  \includegraphics[width=0.8\linewidth]{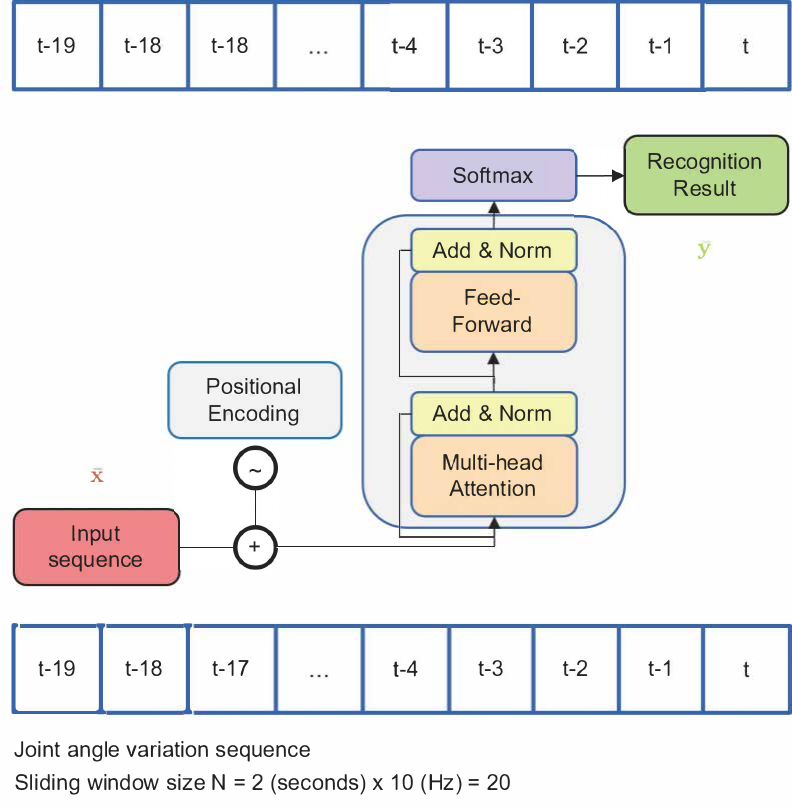}
  \caption{The Transformer takes as input joint angle sequences extracted from a sliding window, and outputs recognition result at each time step.}
  \label{fig:transformer}
\end{figure}
The Transformer \cite{vaswani2017attention} models temporal dependencies in time series via self-attention, making it suitable for human joint trajectories. At each time $t$, it classifies the current action from the previous $N$ joint angle vectors $\bar{\mathbf{q}} \in \mathbb{R}^{22 \times N}$. Each vector is linearly projected and combined with a learnable positional encoding:
\begin{align}
\mathbf{x}_i = \text{PE}(\mathbf{W}_i \mathbf{q}i), \quad i=t-N+1,\dots, t
\end{align}
The sequence $\bar{\mathbf{x}}$ is processed by $L$ Transformer encoder layers, each applying multi-head self-attention and a feed-forward network with residual and normalization layers:
\begin{align}
    \bar{\mathbf{z}}^{(l)} = \text{LayerNorm}(\bar{\mathbf{x}}^{(l-1)} + \text{MultiHead}(\bar{\mathbf{x}}^{(l-1)}) 
    \end{align}
    \begin{align}
    \bar{\mathbf{x}}^{(l)} = \text{LayerNorm}(\bar{\mathbf{z}}^{(l)} + \text{FFN}(\bar{\mathbf{z}}^{(l)}), \quad l=1,\dots, L 
\end{align}
The multi-head attention computes self-attention between all positions in the sequence:
\begin{align}
\label{eq:attention_equation}
    \text{Attention}(\mathbf{Q}, \mathbf{K}, \mathbf{V}) = \text{softmax}\left(\frac{\mathbf{Q}\mathbf{K}^T}{\sqrt{d_k}}\right)\mathbf{V}
\end{align}
where $\mathbf{Q}$, $\mathbf{K}$ , $\mathbf{V}$ are the query, key, and value matrices, representing attention calculation parameters. $d_k$ is the dimension of the key matrix.

Finally, an MLP head maps the last encoder output to $C=17$ action classes:
\begin{align}
    \bar{\mathbf{y}} = \arg \max (\text{softmax}(\mathbf{W}_{\text{MLP}}\bar{\mathbf{x}}^{(L)}))
\end{align}
where $\mathbf{W}_{\text{MLP}} \in \mathbb{R}^{C \times d}$ is the learnable input projection of the MLP head.
Figure~\ref{fig:transformer} illustrates the sliding-window input used for per-frame action recognition.

\subsection{Temporal causal mask}
Causal (autoregressive) masking constrains a Transformer’s self-attention such that each time step attends only to current and past inputs, preventing access to future information \cite{vaswani2017attention}. This constraint is necessary for time-dependent tasks like real-time action recognition, where temporal causality must be preserved. Practically, it is implemented as a lower triangular attention mask, ensuring that future tokens are assigned zero attention weights and thus excluded from the computation.

Given a sequence of length $T$, the causal mask $\mathbf{M} \in \mathbb{R}^{T \times T}$ is defined as:

\begin{align}
\mathbf{M}_{i,j} = 
\begin{cases}
1, & \text{if } j \leq i \\
0, & \text{if } j > i
\end{cases}
\end{align}

This mask is added to the attention logits in \eqref{eq:attention_equation}, before the softmax operation to prevent each position $i$ from attending to any future positions $j > i$.

\subsection{ Loss function}
In the work of Mazzia \cite{mazzia2022action}, a single classification token is assigned to represent the action of an entire sequence. In contrast, our training strategy involves assigning labels to each time step, which encourages the model to learn a more fine-grained and temporally aware representation of actions across the sequence. This dense supervision helps the model better capture the dynamic progression of actions and ultimately improves recognition accuracy at the final time step. Accordingly, in our Transformer-based model, we apply a standard classification loss independently at every time step:
\begin{align}
\label{eq:classfication_loss}
    \mathcal{L}_{cls} = \frac{1}{T} \sum_t - \log (y_{t,c}),
\end{align}
enabling the network to learn temporal dependencies while maintaining strong per-frame prediction capabilities.

The Transformer produces an action label at each time step. However, since predictions are generated independently from the latent representations $\bar{\mathbf{x}}$, no explicit constraint enforces temporal consistency across the output sequence $\bar{\mathbf{y}}$. Consequently, recognition outputs may fluctuate between consecutive time steps. To mitigate this effect, temporal smoothing is introduced to promote consistent predictions over time.

Following the approach of Farha and Gall \cite{farha2019ms}, a combination of classification and smoothing losses is employed during training. The temporal smoothing loss, formulated as a truncated mean squared error (T-MSE), regularizes abrupt variations in consecutive predictions:
\begin{align}
\mathcal{L}_{T\text{-}MSE} = \frac{1}{NC}\sum_{t,c} \tilde{\Delta}_{t,c}^{2},
\end{align}
\begin{align}
\tilde{\Delta}_{t,c} =
\begin{cases}
\Delta_{t,c}, & \Delta_{t,c} \leq \tau, \\ 
\tau & \text{otherwise},
\end{cases}
\end{align}
\begin{align}
\Delta_{t,c} = |\log y_{t,c} - \log y_{t-1,c}|,
\end{align}
where  $y_{t,c}$ the predicted probability of class $c$ at time $t$, and $y_{t-1,c}$ the corresponding probability at the preceding step (excluded from gradient computation), $\tau$ denotes the threshold on consecutive log-probability changes, set to 4 (roughly 50 percent change in probability) to limit abrupt variations while allowing transitions between actions.

The smoothing term is combined with the standard cross-entropy classification loss in \ref{eq:classfication_loss}
resulting in the final training objective:
\begin{align}
\mathcal{L}_{total} = \mathcal{L}_{cls} + \lambda \mathcal{L}_{T\text{-}MSE},
\end{align}
where $\lambda$ controls the relative contribution of temporal regularization, set to $\lambda = 0.15$ in this work.

\subsection{Implementation details}
An advantage of using joint angles as input is that, when action labels are closely linked to specific limb movements, recognition can be performed using only the joint angles of the corresponding limbs. This characteristic aligns well with the organization of our dataset. Accordingly, we use 12 joint angles from the trunk and lower limbs to classify motion labels and their transitions, and 18 joint angles from the trunk and upper limbs to classify ordering action labels and background actions.
Given the size of our dataset, a Transformer model with approximately 20k parameters is considered suitably scaled. Our network is inspired by the work of Mazzia et al. \cite{mazzia2022action}, where the authors proposed four sets of hyperparameters to construct Transformer networks of varying sizes: \textit{AcT-$\mu$}, \textit{AcT-S}, \textit{AcT-M}, and \textit{AcT-L}. The first model, \textit{AcT-$\mu$}, has approximately 20k parameters, while the others have much larger parameter counts and require a substantial amount of training data, making them unsuitable for our dataset. Consequently, \textit{AcT-$\mu$} was chosen as the baseline for comparison.
Moreover, to ensure a fair comparison, we adopt for our model a design similar to \textit{AcT-$\mu$}, using 4 Transformer encoder layers, 1 self-attention head, a model dimension of $d=64$, and an MLP head dimension of 256. The Transformer was trained using the Adam optimizer with an initial learning rate of $10^{-4}$ and a cosine annealing schedule, over 8000 epochs — after which the training loss plateaued and showed minimal further improvement. During testing, the inference time was measured using the previously described hardware settings. 
\subsection{Evaluation metrics}
 To evaluate the effectiveness of our model in comparison to the AcT-$\mu$ baseline, we used the four metrics, shown in the Table \ref{tab:metrics}: accuracy (acc), mean average precision (mAP), macro-averaged F1 score (F1), binary accuracy (binary acc), and processing speed quantified in frames per second (FPS).

The evaluation begins by applying the same model architecture, AcT-$\mu$, to different input features. To assess the model's generalization to viewpoint changes, we apply four distinct and random spatial transformations to the test set JCP data, simulating pseudo-viewpoint shifts. For comparison, we use three types of input: 3D JCPs obtained from the RTMPose pose estimator and triangulation, 3D JCPs generated through LSTM-based augmentation, and joint angles calculated from our pipeline (see Section \ref{pipeIK}). In the case of joint angles, the input is computed from the transformed JCPs, ensuring that all modalities are evaluated under the same viewpoint variations.

A second set of results is reported from ablation studies on the variations of the Transformer architecture, using joint angle inputs. Our newly derived model is named AcT-$\mu$-SAR (for smoothed and auto-regressive).

\begin{table}[t]
\centering
\small
\caption{Different metrics for model evaluation.}
\label{tab:metrics}
\begin{tabular}{lp{5.5cm}}  
\toprule
\textbf{Metric} & \textbf{Definition} \\
\midrule
acc\cite{opitz2024closer}  & 
$\text{acc} = \frac{1}{N}\sum^{N}_{i=1} \mathbf{1}(y_i = \hat{y}_i)$ \\

mAP\cite{opitz2024closer} & 
$\text{mAP} = \frac{1}{C}\sum_{c=1}^{C} \text{AP}(c)$ \\

F1\cite{opitz2024closer} & 
$\text{F1} = \frac{1}{C}\sum^{C}_{c=1} \frac{2 P_c R_c}{P_c + R_c}$ \\

binary acc\cite{opitz2024closer} &  
$\text{binary acc} = \frac{1}{NL} \sum \mathbf{1}(\hat{y} = y)$
 \\
FPS & Frames Per Second \\

\bottomrule
\end{tabular}
\end{table}

\section{Results}
\subsection{Online Human Action Recognition}
\label{Deep_exp}
\begin{figure}[h!]
    \centering
    \includegraphics[width=\linewidth,clip,trim=0 0 0 0]{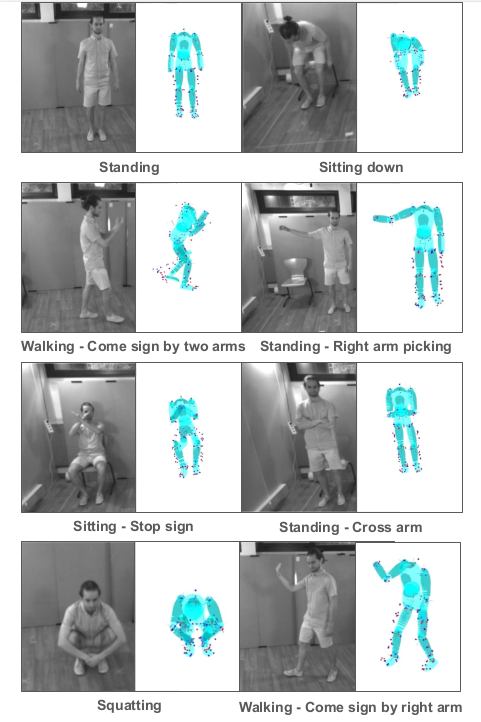}
   \caption{Eight representative tasks shown using one camera view and their corresponding inverse kinematics reconstruction.}
   \label{fig:fig_tasks_ik}
\end{figure}

Fig.~\ref{fig:fig_tasks_ik} shows 8 actions and their corresponding reconstruction using the proposed inverse kinematics pipeline. The proposed pipeline is able to accurately reproduce complex motions, including those involving self-occlusions, e.g., crossing arms or performing a deep squat. This is also visible in the accompanying video (also available at \url{https://bit.ly/humar-ohar-demo}), which shows the joint angle estimation of a subject during one complete trial.

The inference time of the proposed model was 2.77 ms, fulfilling the real-time processing requirement. 

Table~\ref{tab:comparison_feature} compares the performance of identical Transformer models trained with a time-step-wise classification loss across different input feature types. Models using Cartesian-space inputs show lower accuracy due to their sensitivity to viewpoint changes, as the training data does not include spatial transformations of joint center positions (JCPs) as in test set. In contrast, joint angle kinematics provide viewpoint-invariant representations, resulting in more robust action recognition. The reported values illustrate this trend rather than absolute performance. Since only four random spatial transformations were applied to JCPs, the results reflect this limited variability; additional transformations would likely alter the numerical values. Transformer models relying on Cartesian-space inputs perform reliably only when test conditions closely match training data. Achieving robustness to viewpoint changes would require extensive data augmentation over diverse transformations, substantially increasing computational cost.

Table~\ref{tab:ablation_model} summarizes ablation results for the Transformer using joint angle kinematic inputs. The baseline model follows \cite{mazzia2022action}, employing a single classification token for sequence-level recognition. Introducing time-step-level classification with a combined loss improves accuracy, while adding causal masking further enhances performance. The final AcT-$\mu$-SAR integrates both causal masking and the combined loss, achieving the highest accuracy among all variants.

\begin{table}[h]
\centering
\caption{Comparison of input feature types for evaluating model generalization to viewpoint changes}
\label{tab:comparison_feature}
\scriptsize
\begin{tabular}{lccccc}
\toprule
\textbf{\quad \quad \quad \quad Model} & \textbf{acc} & \textbf{mAP} & \textbf{F1} & \textbf{binary acc} \\
\midrule
\makecell{AcT-$\mu$-SAR\\(3D triangulated JCPs)} & 0.4892 & 0.4407 & 0.5921 & 0.7292 \\
\makecell{AcT-$\mu$-SAR\\(3D anatomical markers)} & 0.4647 & 0.4178 & 0.5610 &  0.7380\\
\makecell{AcT-$\mu$-SAR\\(Joint angles)} & 0.8710 & 0.7842 & 0.8757 & 0.9435 \\
\bottomrule
\end{tabular}
\end{table}

\begin{table}[h]
\centering
\caption{Ablation studies on Act-$\mu$ model architecture}
\label{tab:ablation_model}
\scriptsize
\begin{tabular}{lccccc}
\toprule
\textbf{ \quad \quad \quad \quad Model} & \textbf{acc} & \textbf{mAP} & \textbf{F1} & \textbf{binary acc} & \textbf{FPS}\\
\midrule
\makecell{AcT-$\mu$\\(baseline model)} & 0.8491 & 0.7766 & 0.8700 & 0.9418 & 436\\
\makecell{AcT-$\mu$\\(modified loss)} & 0.8645 & 0.7893 & 0.8780 & 0.9461 & 414 \\
\makecell{AcT-$\mu$\\(causal mask)} & 0.8666 & 0.8016 & 0.8860 & 0.9488 & 340 \\
\makecell{AcT-$\mu$-SAR\\(modified loss +\\ causal mask)} & \textbf{0.8810} & \textbf{0.8085} & \textbf{0.8905} & \textbf{0.9510} & 309\\
\bottomrule
\end{tabular}
\end{table}

Our dataset is imbalanced, with transitory actions underrepresented and harder to classify. This was expected as they last shorter than actual actions. This is the case, for example, when standing up from a chair. Also, since the dataset is annotated with multiple labels from four category sets, at each time step, the model predicts both a lower-limb label and an upper-limb label. Table \ref{tab:f1_per_class} shows the recognition results for each category. Nevertheless, the recognition of transitory motions is less important in the current setting aiming at controlling a robot. 
\begin{table}[t]
\centering
\scriptsize
\caption{average F1 scores per class category.}
\label{tab:f1_per_class}
\begin{tabular}{cccccc}
\toprule
\textbf{Motions} & \textbf{Transitory actions} & \textbf{Ordering actions} & \textbf{Background actions}  \\
0.9779 & 0.7572& 0.8900 & 0.8783 \\
\bottomrule
\end{tabular}
\end{table}

The model performs well on motion categories but struggles with ordering and background actions, which involve upper-limb labels. The absence of hand joint angles makes it difficult to distinguish actions that differ only by subtle variations in elbow movements, such as "come sign with two arms", "stop sign" and "two thumbs up".

These results demonstrate that joint angle kinematics is an effective representation of input data for achieving viewpoint-invariant action recognition. Furthermore, incorporating a classification loss at each time step, along with the application of causal masking and smoothing loss to enforce temporal consistency, enhances the Transformer model’s ability to accurately recognize actions.

\subsection{Experiment in simulation using OHAR for human-robot interaction }
\label{robotexp}
To evaluate the effectiveness of using joint angle kinematics as input to a Transformer-based model for OHAR, we conducted a preliminary experiment in a robotic simulation environment designed to emulate human-robot interaction. The setup integrates the proposed real-time markerless inverse kinematics pipeline and the Transformer-based action recognition model, enabling continuous and reactive motion understanding. 

Recognized human actions are transmitted via a UDP connection to a MuJoCo simulation controlling a KUKA robotic arm equipped with a virtual pen. The robot is driven by two control schemes:

\begin{itemize}
    \item Position control for the pen's vertical motion, determining the pen’s contact with a virtual whiteboard, and
    \item Velocity control task for horizontal motion, constrained within the arm’s reachable workspace.
\end{itemize}

Robot behavior is directly conditioned on the detected human actions, following a set of intuitive mappings:

\begin{itemize}
    \item When "Sitting down” is detected, the pen is lowered to establish contact with the whiteboard.
    \item "Left arm picking”, “right arm picking”, “two arms picking”, and “come sign by two arms” control the pen to move horizontally in the directions right, left, backward, and forward, respectively.
    \item Upon the detection of “Walking”, the pen is lifted to a resting position above the board, and the ink trace is cleared.
\end{itemize}

\begin{figure}[t]
  \centering
  \includegraphics[width=1\linewidth,clip,trim=0.8cm 0 0 0]{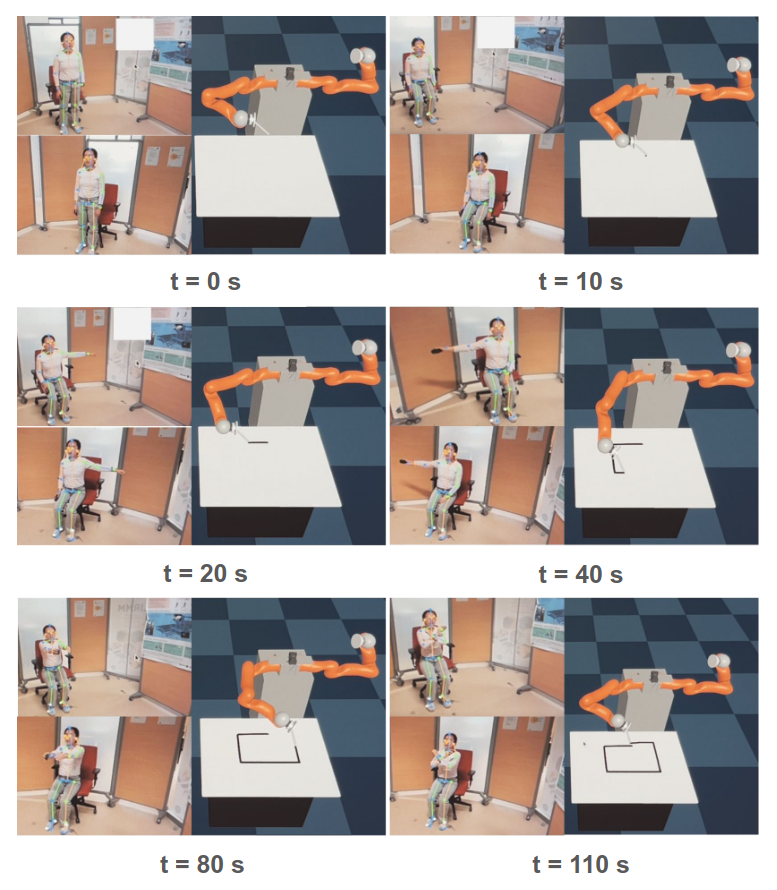}
  \caption{Snapshots of the proposed prelininary experiment, described in the accompagnying video, performed with a simulated robot.
  \newline t=0s: standing, robot INIT state. t=10s: sitting down, robot PEN DOWN state. t=20s: left arm picking, robot PEN RIGHT state (control in mirror). t=40s: right arm picking, robot PEN LEFT state. t=80s: two arms picking, robot PEN BACKWARD state. t=110s: stop sign, robot PEN PAUSE state}
  \label{fig:caligraphy}
\end{figure}

This configuration demonstrates a simplified calligraphy scenario, where a sequence of recognized actions produces a continuous ink trajectory forming a square on the virtual whiteboard (Figure~\ref{fig:caligraphy}). A human subject performs actions from the HUMAR-2024 dataset, captured in real time through the IK estimation pipeline and processed by the Transformer model. The model outputs frame-wise class probabilities, from which the final action label is determined using an argmax operation. To improve robustness, we employ a 2-second temporal buffer, confirming an action when its label occupies more than 50\% of the buffered frames—thus filtering out transient misclassifications at the cost of minor latency.


This experiment validates the feasibility of using joint angle representations for real-time, vision-based human–robot interaction, while highlighting the inherent trade-off between recognition stability and system reactivity. In particular, the 2-second temporal buffer introduces an effective delay of about 1 second, which explains why the square trajectory drawn by the robot does not fully close in the demonstration. Future work will focus on integrating the real-time action recognition pipeline into more industrial-like scenarios, which demand immediate action recognition and contextual understanding of human motions for scene interpretation.

 \section{Conclusion}

In summary, this paper presents an end-to-end, real-time joint kinematics and action recognition pipeline using two affordable cameras. The system achieves a processing time of 47.5 milliseconds when running on limited hardware resources. The estimated joint angles are processed by a series of machine learning models to identify representative actions in a human–robot collaboration context. By using joint angle kinematics as input features, the Transformer model, trained with a time-step-level combined loss and causal masking, achieves an accuracy of $88.1\%$ for real-time human action recognition. A preliminary experiment integrating the action recognition pipeline into a robotic calligraphy scenario further demonstrates the feasibility of using joint angle–based recognition for responsive and interactive human–robot systems in dynamic environments.


\section*{ACKNOWLEDGMENT}
This work was supported by "Défi Clé Robotique Centrée sur l'Humain" funded by Région Occitanie, France, and the ANR HERCULES (ANR-23-CE33-0010).

\bibliographystyle{IEEEtran} 
\balance
\bibliography{biblio.bib}
\addtolength{\textheight}{-4cm}   

\end{document}